\documentclass[conference]{IEEEtran}
\IEEEoverridecommandlockouts
\usepackage{cite}
\usepackage{times}
\usepackage{latexsym}
\usepackage{comment}
\usepackage{algorithm}
\usepackage[noend]{algcompatible}
\usepackage{todonotes}
\usepackage{multirow}
\usepackage{enumitem}
\usepackage{amsmath,amssymb,amsfonts}
\usepackage{graphicx}
\usepackage{textcomp}
\usepackage{xcolor}
\def\BibTeX{{\rm B\kern-.05em{\sc i\kern-.025em b}\kern-.08em
    T\kern-.1667em\lower.7ex\hbox{E}\kern-.125emX}}
    
\begin{document}
\title{Personalization of Deep Learning}
\author{\IEEEauthorblockN{Johannes Schneider}
\IEEEauthorblockA{Institute of Information Systems, \\
University of Liechtenstein \\
Vaduz,Liechtenstein \\
johannes.schneider@uni.li}
\and
\IEEEauthorblockN{Michalis Vlachos}
\IEEEauthorblockA{Department of Information Systems \\
University of Lausanne \\
Lausanne, Switzerland\\
}
}

\maketitle

\begin{abstract}
	 We discuss training techniques, objectives and metrics toward personalization of deep learning models. In machine learning, personalization addresses the goal of a trained model to target a particular individual by optimizing one or more performance metrics, while conforming to certain constraints. To personalize, we investigate three methods of ``curriculum learning`` and two approaches for data grouping, i.e., augmenting the data of an individual by adding similar data identified with an auto-encoder. We show that both ``curriculuum learning'' and  ``personalized'' data augmentation lead to improved performance on data of an individual. Mostly, this comes at the cost of reduced performance on a more general, broader dataset. 
\end{abstract}
\section{Introduction} \label{sec:intro} 
Personalization is a well-established and important topic in computer science and cognitive sciences \cite{fan06} with many applications, e.g., in recommender systems \cite{zha18} and personal assistants \cite{Sar17}. It also has numerous applications in medicine, eg. \cite{lee2019,fink17,gol19}. Personalization of machine learning models may have multiple, partially conflicting objectives, for example, optimizing both performance (``How well does the system perform with regards to data of the individual and in general?'') and non-task related measures such as privacy and fairness. As an example, a speech or character recognition system of a personal assistant might primarily classify data of a single user. So it is more important to "personalize" and recognize characters of that particular user better than from other users.

Here, we elaborate on personalization in deep learning, presenting a) learning techniques, b) objectives and c) evaluation metrics.  Personalization is accomplished by using a different (training) dataset for each individual in combination with a larger global dataset shared among all individuals. We assume that the global training data provides valuable information beyond the individual's preferred outputs. 

\begin{figure}[!ht]

	\centering
	\includegraphics[width=0.5\textwidth]{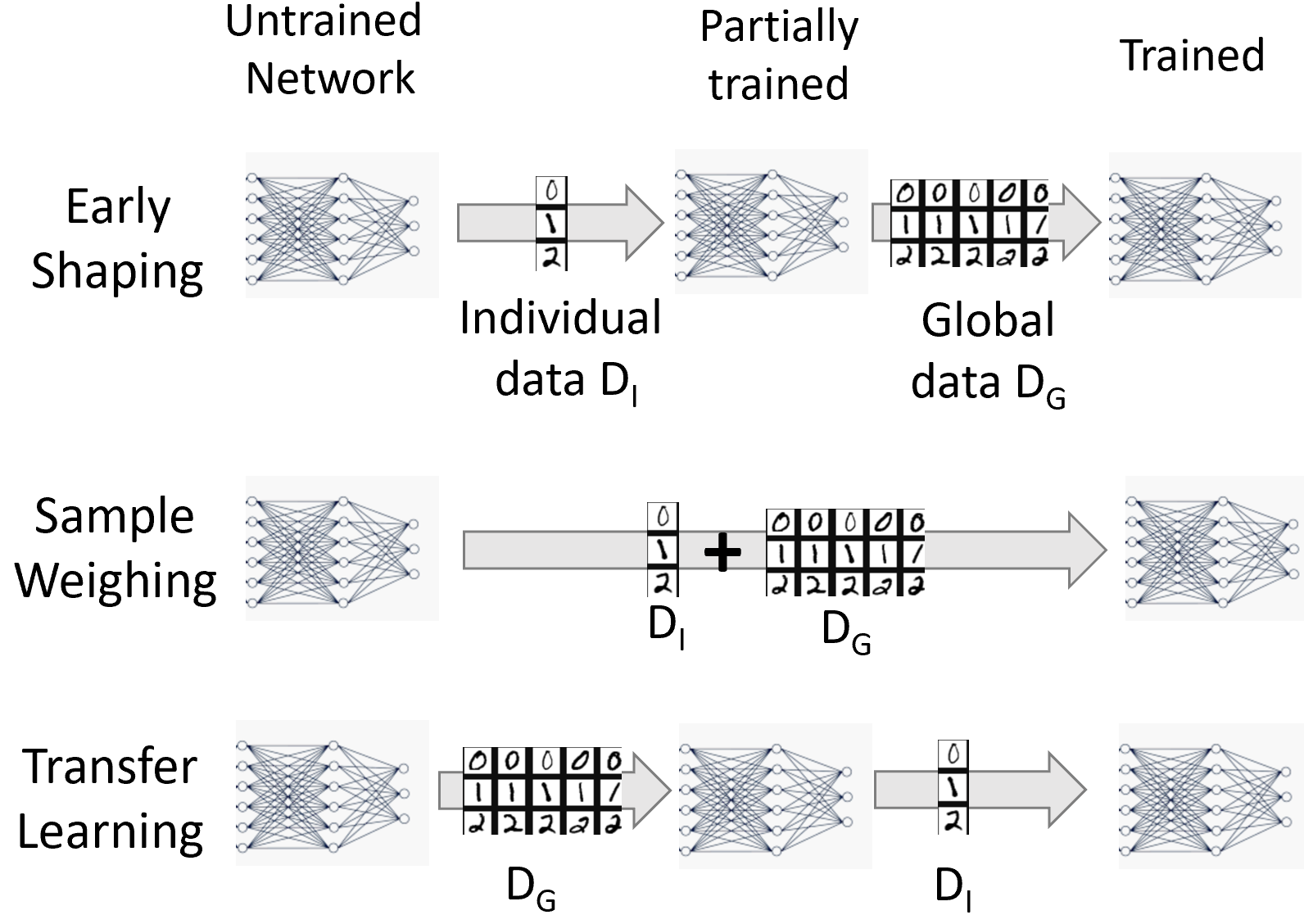}
	\caption{The three forms of shaping used in this work} \label{fig:shaping3Overview}
\end{figure}

Our methods are influenced by two concepts or ideas: \textit{shaping} and \textit{data grouping}. Shaping, in the context of psychology~\cite{pet04}, is the process of strengthening a behavior through punishment and reinforcement. In our case, we shape the features of a network. That is, we aim at personalization by increased exposure (or weighing) of data of an individual. For a neural network we perform ``early'' (or ``late'') shaping, by training a network in the initial (or final) phases of training exclusively with data of an individual. Late shaping is also known as transfer learning~\cite{ben12} or fine-tuning. We also investigate a form of sample weighing by intertwining training on data from the individual and (global) training data. The three forms of shaping, illustrated in Figure \ref{fig:shaping3Overview}, can also be combined. Data grouping refers to the idea of augmenting the dataset of an individual (which is often small), by adding similar samples found in the global dataset. To do so, we train an auto-encoder to find a latent representation of the input space. We then use the Euclidean distance of latent representations of samples to compute similarity among samples.


\begin{figure}[!ht]
	\centering
\includegraphics[width=0.46\textwidth]{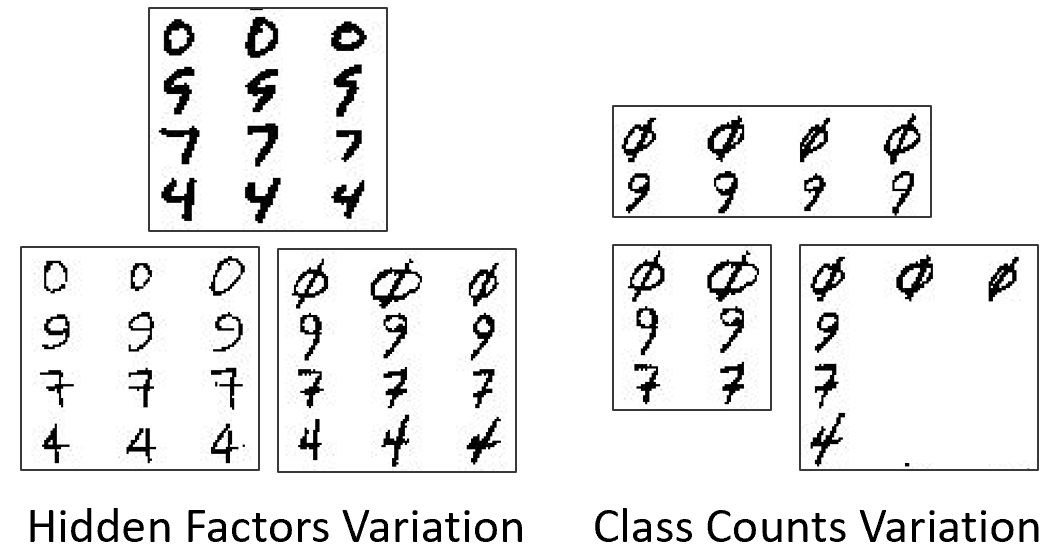}
 	\caption{A box shows data of individuals varying in class counts or hidden factors} \label{fig:var}
\end{figure}

\section{Definitions and Problem Description} \label{sec:defs}
We focus on classification of data for a set $S$ of individuals. Each individual $I \in S$ provides a dataset $D_I$. The set of all $|S|$ datasets is $D_S:=\{ D_I | I \in S \}$. A dataset $D_I$ of an individual $I$ consists of inputs $X$ and labels $y$, ie. $D_I:=\{(X,y)\}$.  Additionally, we assume that we are given a general dataset $D_G$ that can be used for training. This dataset may be the union of datasets from individuals, ie. $D_G = D_S$. However, generally due to privacy reasons, it cannot be assumed that the dataset of each individual can also be used for the training of models of other users. Furthermore, the data from individuals may be biased or it may not be of large quantity or high quality, eg. contain many mis-classifications, and, thus, additional data processing or data from other sources will be required. We assume that input samples $X$ from each $D_I$ and $D_G$ are homogeneous, ie. drawn from the same space. We assume that both contain the same classes $C$, but some datasets in $D_S$ might not contain samples from all classes $C$. Each dataset $D \in (D_S \cup D_G)$ is further partitioned into a training, validation and test data set. We do not introduce explicit notation to refer to these subsets, but explicitly state the subset, eg. validation data from $D_I$, if there might be doubt to what $D_I$ or $D_G$ might refer to. The sizes of the sets $D_G$ and $D_I$ are unconstrained. Our methods are tailored to the scenario, where $D_G$ is significantly larger than any dataset $D_I$. Labels of the individual dataset $D_I$ might be conflicting with those in $D_G$ or from other individuals $D_{I'}$ -- even for similar examples. While exact duplicates of samples in $D_I$ are not likely to occur in data of other individuals or in $D_G$, homogeneity implies that for the majority of individuals $I \in S$ the global dataset $D_G$ might contain similar samples to those of the individual $I$.

We consider two independent variants how samples differ among individuals, illustrated in Figure \ref{fig:var}:
\smallskip
\begin{enumerate}[label= (\arabic*) ]
	\item \textbf{Class counts variation}: The distribution of the number of samples per class is different among individuals. The rational is that if an individual shows particular interest, expertise or preference for one class, she should also have had more exposure to samples of that class. In a recommendation setting, for instance, a person that prefers action movies, might have watched more action movies than people preferring dramas. This might also be observed as a consequence of growing up at a particular location, eg. people in some regions of the world might be very familiar with many different kinds of snow and deer while in other areas they are more familiar with different kinds of sand and fish.
	\smallskip
	\item \textbf{Hidden factors variation}: There is considerable variation of samples within classes per individual that can be partially explained by hidden factors per class and, possibly, per individual. We assume that samples of a class from $D_I$ are on average more similar to each other than samples from $D_G$. The similarity metric as well as factors characterizing similarity might be latent. In a recommendation setting, for instance, two persons might have watched the same number of action movies and dramas, but one of the persons prefers male main actors and the other female main actors. Differences due to growing up might manifest similarly, eg. two people have roughly seen the same number of dogs and deer, but a person from Sweden might have seen more huskies and a person from Austria more shepherd dogs. All shepherd dogs share commonalities that differ from huskies, but both huskies and shepherd dogs belong to the same class, ie. dogs. The two breeds of dogs can be seen as sub-classes of dogs. Thus, generally, it can be said that each individual is exposed to a different set of sub-classes characterized by unknown latent factors.
\end{enumerate}

With respect to the training of deep learning models we use standard conventions but require a few clarifications of terms. We assume that the network parameters are altered using gradient descent based on batches, ie. a batch is a small set of randomly chosen samples. In our algorithms, all samples of a batch are chosen entirely from either $D_I$ or $D_G$.

\subsection{Objectives and Metrics of Personalization} \label{sec:mass}
The purpose of personalization is to obtain a model tailored to each individual. While we primarily focus on quantifying a model's utility using performance-oriented metrics, we also briefly address fairness as an example of a metric in the field of legal and ethical concerns. Other metrics related to computational and storage resources, privacy and interpretability are not investigated here (the interested reader is refered to \cite{sho15,aba16,sch19}). 



\begin{enumerate}[label=(\arabic*)]
		
\item\textbf{Personalized-data performance}: The system should derive the same decisions as specified by the individual. This holds, potentially, even if these decisions are non-optimal with respect to performance metrics on a global dataset, serving as ground truth. That is to say, that an individual's training data might differ strongly from the global training data or of data of other individuals, eg. since the individual systematically misclassifies examples for one class. As metric for our classification task, we use the classification accuracy $acc(D_I)$ on a test set of $D_I$. The accuracy on the entire population of all individuals, ie. the datasets in $D_S$ is the mean of the accuracy of all datasets, ie. $acc(D_S):= \sum_{I \in S} acc(D_I) /|S|$.  
\smallskip
\item\textbf{Global-data performance}: This refers to performance for the task at hand in general as judged objectively rather than based on data from an individual. That is, data of any individual is deemed irrelevant in computing the task performance. A large global dataset $D_G$ that could be the consensus of multiple people or it could consist of objective and accurate measurements. 
		As metric, we use the classification accuracy $acc(D_G)$.
	
\smallskip
\item\textbf{Fairness}:  Fairness refers to the degree to which treatment is egalitarian \cite{bin17,kus17}. There are many notions of fairness based on different assumptions and intentions\cite{kus17}. We focus on individual fairness \cite{dwo12} that aims at treating similar individuals equivalently. Computing similarity among individuals requires a meaningful similarity metric, which by itself makes ethical statements by answering questions such as ``What matters more for similarity: Race or gender?'' Similarity metrics (in general) are provided by many papers, eg. \cite{kim18f}. We propose the L2-norm in the auto-encoded space(Section \ref{sec:pers}). But, we shall treat all humans equally rather than distinguishing groups of individuals. That is to say, the machine learning system should personalize (close to) equally well on data from each individual. In practice, this means that variances in performance metrics across individuals might only be tolerable up to some extent and primarily if they do not originate from the method, but are inherent due to differences in the training data. For instance, assume data of an individual and the global (shared) data only differ in the decision (ie. both have the same inputs).  Then, adjusting decisions towards data of the individual worsens the performance on the global dataset, since both are in disagreement. Generally, domain expertise might be necessary to judge whether variances between individuals can be considered fair. 
As metric, we advocate that for any performance metric 
individuals, eg. $err(D_S)$, one should also report standard deviation and spread, ie. the best and worst performance of an individual. Using the minimum and maximum is motivated by the fact that legislation has entered the field of AI through the GDPR, such as the "right to an explanation". Since it grants rights to individuals, the method should also deliver adequate explanations to all individuals. Thus, the worst performance observed in the dataset should still be compliant with regulation. Large spread is an indicator of low fairness. Furthermore, minimum, maximum and standard deviation are also interesting measures to analyze the performance of the system beyond fairness. 
\end{enumerate}

The above objectives might be conflicting and require trade-offs, eg. between task-data and personalized-data performance. In fact, it seems possible that the machine learning model is encouraged to learn or adjust representations that overfit to the data on the individual. For privacy and fairness, in a commercial setting the goal might be to simply meet basic minimum requirements in order not to be in conflict with the law. Both privacy and fairness aspects might negatively impact task performance.

Here, we focus on the goal to maximize classification accuracy on one's dataset $D_I$, while ensuring a minimum performance level on the general dataset $D_G$. Additionally, we aim to fulfill constraints related to other objectives. 

\section{Personalization Methods} \label{sec:pers}
Our approach is to alter the outcome of the training process of a deep learning model through curriculum learning using three forms of shaping as well as by data grouping, ie. enhancing the dataset of an individual with similar data. For shaping, we change the frequency (and thereby relevance) of data of an individual used throughout the training process. Due to increased exposure to the data of an individual, the learning process will push the features learnt by the network to resemble more closely patterns found in the data of the individual. Our strategies for curriculum learning are illustrated in Figure \ref{fig:shaping3Overview}.  


\subsection{Early shaping} \label{sec:early} 
Early shaping refers to personalizing a deep learning model early in the training process. While it is well-known that for human brains, ie. biological neural networks, experiences in early childhood, in particular extreme forms such as traumas, might have an impact throughout the entire life-time of an individual, in a deep learning context this is less evident. Deep learning suffers from ``catastrophic forgetting'', meaning that learnt knowledge that is still valuable is forgotten in the process of continuous learning, e.g. \cite{kir17}. As such, it is unclear, whether training in later phases might not undo the initial shaping. On the other hand, the success of training relies on initial conditions \cite{sut13}, ie. proper random initialization. Therefore, strong adjustments towards a dataset of individuals might be problematic. They could manifest in overfitting and relatively variation in the learnt representations. This holds in particular, since the dataset $D_I$ of an individual might not cover many relevant factors found in the general dataset $D_G$, increasing the risk for overfitting to $D_I$ that might not be reversible.  In this paper, we adopt an extreme form of early shaping. That is, we train a model using only the data $D_I$ of an individual for a limited timespan before proceeding to train only on $D_G$. Unfortunately, determining the right amount of training on $D_I$ is non-trivial. Training for few epochs might not lead to a significant degree of shaping, ie. later training epochs (using data not from the individual) might undo the small changes done by initial shaping. Training for many epochs appears to be the more viable option if adequate regularization counteracting overfitting is in-place. In such a setup, training of deep learning networks for too long does not show very strong adverse effects. That is to say, overfitting can be a relevant problem, but it is often not a key concern. In contrast, we observed that training for too long on a small data set might lead to effects that might not be reversible. The result might be bad performance as hypothesized earlier in the paragraph. We shall discuss this phenomenon in more detail in our experiments section. \newline
Early shaping and the other two methods are illustrated in Figure \ref{fig:shaping3Overview} and described in Algorithm 1.

\begin{figure}
	\centering
	\includegraphics[width=0.5\textwidth]{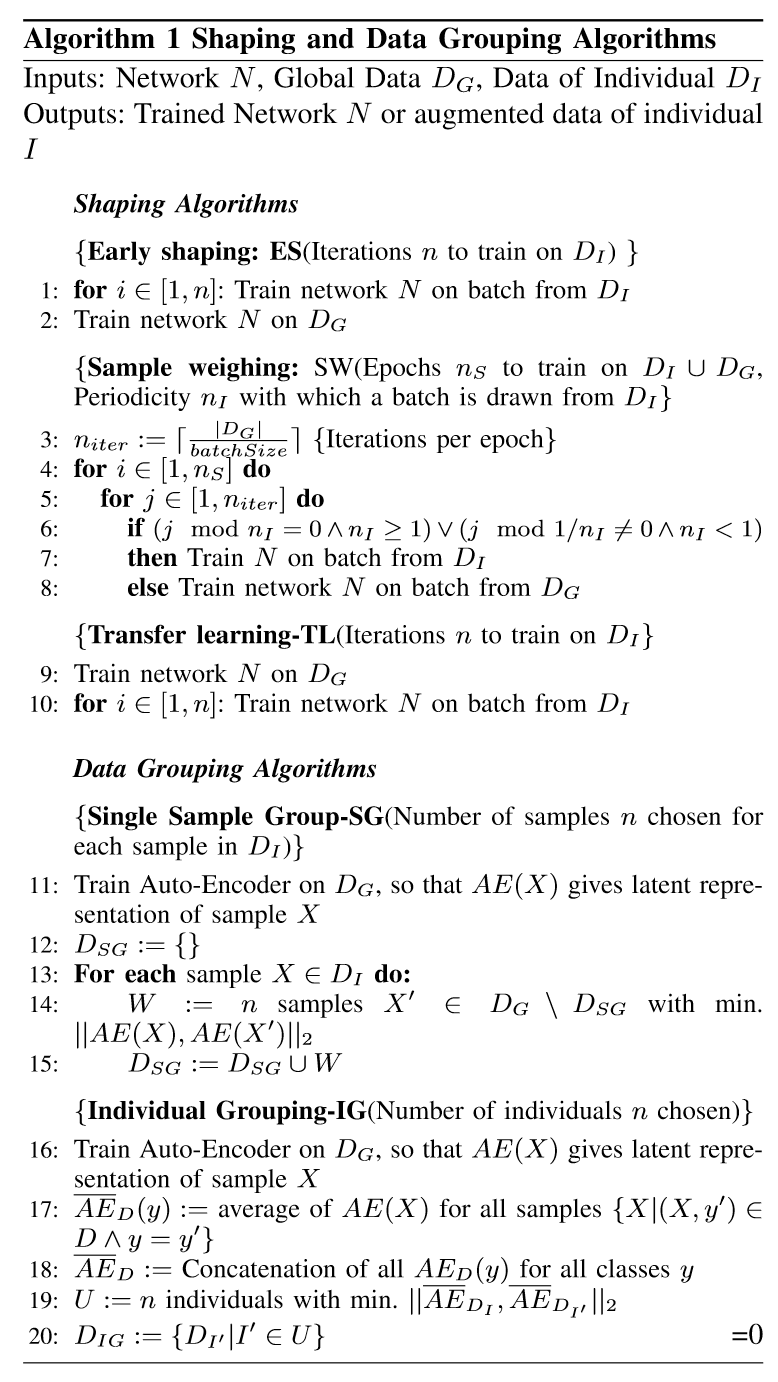}
\end{figure}

\subsection{Sample weighing} Data on the individual should have a strong influence on the model. This can be achieved by more exposure to the data of an individual throughout the entire training process. The idea of sample weighing is to train on the dataset $D_G$ and to take every $n_I$-th (for $n_I>1$) batch from $D_I$ rather than $D_G$, so that samples from $D_I$ occur more frequently than those in $D_G$. The parameter $n_I$ controls the influence of data $D_I$ on the network. We also allow $n_I<1$. In this case, the exposure to $D_I$ is even larger, ie. all but every $1/n_I$-th batch stems from $D_I$. The pseudo-code is given in Algorithm 1. Note that as stated in the algorithm, the number of iterations $n_{iter}$ per epoch is only sufficient to consider each sample in $D_G$ once. Thus, for parameter $n_I=2$ only half the samples in $D_G$ are considered per epoch, while data $D_I$ might be used multiple times for training. 

\subsection{Transfer learning} 
 In late shaping (or transfer learning) a trained model is updated using data from a different distribution. Transfer learning is appealing, since it is potentially faster, ie. it is not necessary to train the model from scratch for each user. A model can be trained for all users, ie. based on the dataset $D_G$. The trained model can be adjusted to each user with a relatively short amount of training using the data $D_I$ of the individual only. As for early shaping, training for too long might have adverse effects, eg. due to ``catastrophic forgetting'' or overfitting. Thus, some care is needed to limit the impact of transfer learning. The influence of the data $D_I$ of an individual is controllable with the number of iterations $n_{TL}$ on the individual data $D_I$ as shown in Algorithm 1 as well as by reducing the number of layers that are adjusted during the learning on $D_I$. 

\begin{figure*}
	\centering
	\includegraphics[width=0.99\textwidth]{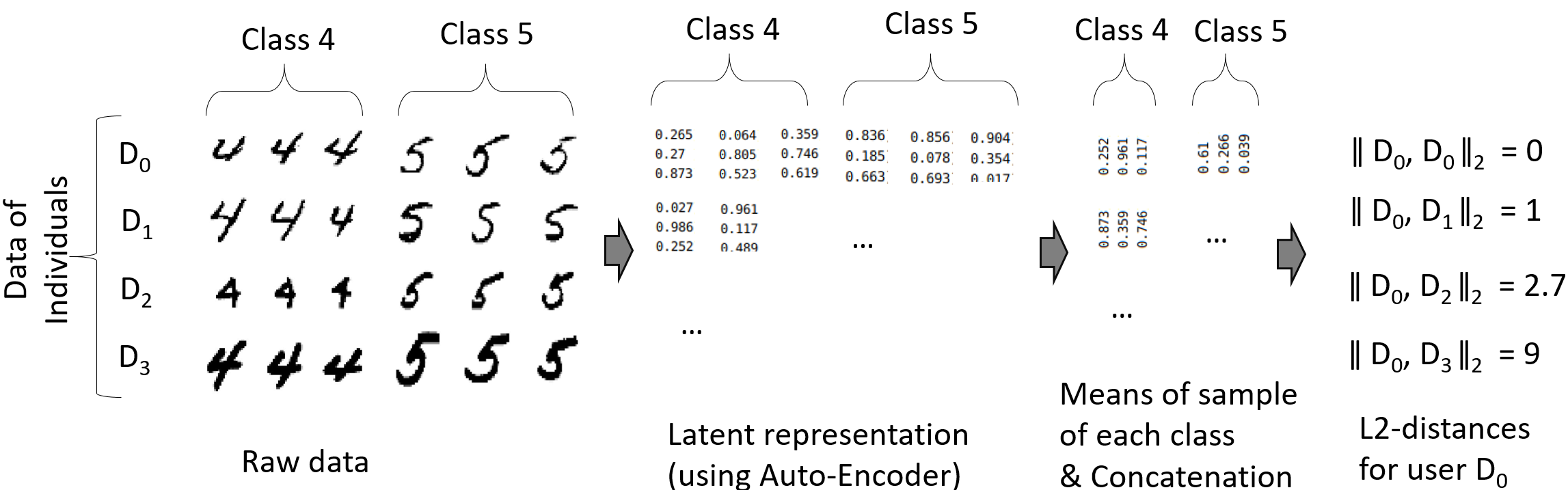}
	\caption{Similarity of data $D_I$ of individual $I$. For dataset $D_0$, $D_1$ is most similar and $D_3$ the least.} \label{fig:ugroup}
\end{figure*}

\textbf{Data grouping:} With data grouping, we aim at enlarging the data of an individual $D_I$ by adding similar data. Then, instead of shaping with the small dataset $D_I$, we utilize the enlarged dataset of the individual $I$. While each individual differs from each other, it can be expected that either entire datasets of individuals are similar or that for a sample there exist similar samples (originating potentially from different individuals). This yields two approaches. Given a dataset $D_I$ for an individual, single sample grouping $SG$  chooses for each sample in $D_I$ the most similar samples in the global dataset $D_G$ (without duplicates), while individual grouping $IG$ identifies datasets of individuals that are similar.
To compute similarity between samples, we use the L2-norm of a latent representation of samples. The representation is computed using the encoder part of an auto-encoder that is trained on all data. To compute similarity between individuals, we encode all samples, compute means of the latent representations per class and compute the L2-norm on the concatenation. The idea is illustrated in Figure \ref{fig:ugroup} and both data grouping algorithms $IG$ and $SG$ are given in Algorithm 1.

\section{Evaluation} \label{sec:eval}
We assess the proposed methods for shaping and data grouping under variation among individual data due to hidden factors but not due to variation of class counts (Figure \ref{fig:var}), since it is arguably the more interesting scenario. We shall evaluate class count variation in future work. We utilize the NIST Special Database 19 that contains handwritten characters of mainly digits\footnote{https://www.nist.gov/srd/nist-special-database-19} of about 4000 users.  We cropped a patch of 32x32 out of the 128x128 images, which was sufficient to cover the digits in their entirety for almost all cases.  A user contributed on average 103 digits. To have more data per user and since character recognition of digits is considered easy, we added the horizontally and vertically flipped images as separate classes, giving a total of 30 classes and about 300 samples per individual $I$ that constitute the data $D_I$. The flipping increases detection difficulty considerably, since some digits such as ``0'' and ``8'' are symmetric. However, in practice, irregularities arise in many forms, i.e. due to tilting, jitter, curvature, start or end of strokes. We observed that they are often fairly consistent for a user (as observable in Figures \ref{fig:var} and \ref{fig:ugroup}). A person is also very familiar with her own handwriting and her personal assistant might primarily focus on detecting her handwriting. As such it seems reasonable to use this data to personalize.  We employed 100 users, ie. $|S|=100$, and used additionally data from 100 users, ie. about 30000 samples, not in any dataset $D_I \in D_S$, ie. $|D_G|=30000$. When personalizing for user $I$ we utilized 70$\%$ of her data $D_I$ for training and the rest for testing.
\begin{figure}
        \centering
		\footnotesize{		
		\begin{tabular}{| l | l|| l | l| }\hline
			Type/Stride& Filter Shape & Type/Stride& Filter Shape \\ \hline
			Conv/s1     & $3\tiny{\times} 3 \tiny{\times} 1 \tiny{\times} 32$  &Continued...   & Continued...\\ \hline
			Conv/s1     & $3\tiny{\times} 3 \tiny{\times} 32 \tiny{\times} 64$ &Conv/s1     & $1\tiny{\times} 1 \tiny{\times} 128 \tiny{\times} 128$\\ \hline 			
			MaxPool/s2 & Pool $2 \tiny{\times} 2$  &MaxPool/s2 & Pool $2 \tiny{\times} 2$\\ \hline
			Dropout/s1;p=.2 &  $64 \tiny{\times} 1$ &Conv/s1     & $1\tiny{\times} 1 \tiny{\times} 128 \tiny{\times} 256$ \\ \hline
			Conv/s1     & $3\tiny{\times} 3 \tiny{\times} 64 \tiny{\times} 128$&MaxPool/s2 & Pool $2 \tiny{\times} 2$\\ \hline
			Conv/s1     & $3\tiny{\times} 3 \tiny{\times} 128 \tiny{\times} 128$&Dro./s1;p=.35 &  $256 \tiny{\times} 1$ \\ \hline
			MaxPool/s2 & Pool $2 \tiny{\times} 2$&FC/s1 & $256 \tiny{\times}$ nClasses \\ \hline
			Conv/s1     & $3\tiny{\times} 3 \tiny{\times} 128 \tiny{\times} 128$&Softmax/s1 & Classifier \\ \hline
		\end{tabular}}
	\caption{VGG-style network architecture \cite{sim14}}  \label{tab:arch} 
\end{figure}

\begin{figure}
        \centering
        \footnotesize{		
        		\begin{tabular}{| l | l|| l | l| }\hline
			Type/Stride& Filter Shape & Type/Stride& Filter Shape \\ \hline
			Conv/s1     & $3\tiny{\times} 3 \tiny{\times} 1 \tiny{\times} 16$  &Continued...   & Continued...\\ \hline
			MaxPool/s2 & Pool $2 \tiny{\times} 2$  &Dense & 512\\ \hline
			Conv/s1     & $3\tiny{\times} 3 \tiny{\times} 16 \tiny{\times} 32$ &Reshape     & $2\tiny{\times} 2 \tiny{\times} 128$\\ \hline 			
			MaxPool/s2 & Pool $2 \tiny{\times} 2$  &Conv/s1     & $3\tiny{\times} 3 \tiny{\times} 128 \tiny{\times} 128$\\ \hline
			Conv/s1     & $3\tiny{\times} 3 \tiny{\times} 32 \tiny{\times} 64$ &NN-Upsample & $2\times 2$\\ \hline 			
			MaxPool/s2 & Pool $2 \tiny{\times} 2$  &Conv/s1     & $3\tiny{\times} 3 \tiny{\times} 64 \tiny{\times} 64$\\ \hline
			Conv/s1     & $3\tiny{\times} 3 \tiny{\times} 64 \tiny{\times} 128$&NN-Upsample & $2\times 2$\\ \hline
			MaxPool/s2 & Pool $2 \tiny{\times} 2$  &Conv/s1     & $3\tiny{\times} 3 \tiny{\times} 32 \tiny{\times} 32$\\ \hline
			  Conv/s1     & $3\tiny{\times} 3 \tiny{\times} 128 \tiny{\times} 128$ &NN-Upsample & $2\times 2$\\ \hline
			 	MaxPool/s2 & Pool $2 \tiny{\times} 2$   &Conv/s1     & $3\tiny{\times} 3 \tiny{\times} 16 \tiny{\times} 16$\\ \hline
			Flatten&& Conv/s1     & $3\tiny{\times} 3 \tiny{\times} 16 \tiny{\times} 1$\\ \hline
			
			Dense & $16$&&\\ \hline
		\end{tabular}}
       \caption{Auto-encoder architecture} \label{fig:aearch}
\end{figure}

We used a VGG-style network \cite{sim14} depicted in Table \ref{tab:arch}  as well as the auto-encoder shown in Figure \ref{fig:aearch}. The auto-encoder also uses a ReLU activation function followed by a batch-normalization layer after each convolutional layer, except the last one, which is only followed by a ReLU layer. The decoder uses nearest neighbor upsampling (NN-upsample). The encoder compresses the input image to a 16 dimensional latent space that we use for similarity computation. The auto-encoder minimizes an L2-reconstruction loss, ie. $(X-X')^2$, where $X$ is the original image and $X'$ the image that arose due to encoding and decoding of $X$ by the auto-encoder. Both the auto-encoder and the VGG network are trained using the Adam-Optimizer for 20 epochs on $D_G$ (and $D_I$ for sample weighing) and a batch size of 128.\footnote{We found the number of epochs to be sufficient to ensure convergence.}  We trained one model per parameter configurations, yielding more than 1000 trained models. For shaping, the baseline method was trained on the union of $D_G$ and data of one individual $D_I$. Adding an individual is necessary so that all shaping methods and the baseline method have the same number of different training samples. For the baseline the evaluation was carried out using data from another individual $D_{I'}$, ie. $D_I \neq D_{I'}$.\footnote{Otherwise the baseline would be equivalent to our sample weighing method.}  For data grouping, we replace $D_I$ by an extended dataset $D'_I \in \{D_{IG},D_{SG}\}$ originating from one of the two data grouping algorithms for training but still test on the validation data of the original $D_I$. We used sample weighing with $n_I=2$, meaning that for every second iteration a sample from the extended dataset of $D_I$ was chosen. For the baseline, we chose 1500 random samples as dataset $D'_I$. 

\begin{table*}[!htb]
	\centering
	\begin{tabular}{|c|c|c| c|c|c| c|c|c| c|c| c|		} \hline
	 \multicolumn{2}{|c|}{\multirow{2}{*}{Method}}&\multicolumn{2}{|c|}{Perso. Perf.}&\multicolumn{3}{|c|}{Fairness(Pers.)}&\multicolumn{2}{|c|}{Task Perf.}&\multicolumn{3}{|c|}{Fairness(Task)}\\ \cline{3-12}
	\multicolumn{2}{|c|}{}&Acc. on $D_I$& Rank&sd&min&max&Acc. on $D_G$&Rank&sd&min&max\\ \hline
	\multirow{7}{*}{\rotatebox[origin=c]{90}{\parbox[c]{1cm}{\centering Shaping}}}	
&BaseLine&.98&6&\scriptsize{.027}&\scriptsize{.86}&\scriptsize{1.0}&.979&5&\scriptsize{.002}&\scriptsize{.971}&\scriptsize{.983}\\\cline{2-12}
&EarlyShape-ES-$n$-40&.981&5&\scriptsize{.031}&\scriptsize{.82}&\scriptsize{1.0}&.979&4&\scriptsize{.002}&\scriptsize{.971}&\scriptsize{.982}\\\cline{2-12}
&EarlyShape-ES-$n$-400&.983&3&\scriptsize{.027}&\scriptsize{.859}&\scriptsize{1.0}&.98&1&\scriptsize{.001}&\scriptsize{.977}&\scriptsize{.982}\\\cline{2-12}
&SampleWeigh-SW-$n$-12&.98&7&\scriptsize{.021}&\scriptsize{.904}&\scriptsize{1.0}&.971&6&\scriptsize{.007}&\scriptsize{.951}&\scriptsize{.981}\\\cline{2-12}
&SampleWeigh-SW-$n$-2&.984&1&\scriptsize{.017}&\scriptsize{.92}&\scriptsize{1.0}&.968&7&\scriptsize{.009}&\scriptsize{.946}&\scriptsize{.979}\\\cline{2-12}
&TransferL-TL-$n$-10&.982&4&\scriptsize{.027}&\scriptsize{.851}&\scriptsize{1.0}&.98&2&\scriptsize{.001}&\scriptsize{.974}&\scriptsize{.982}\\\cline{2-12}
&TransferL-TL-$n$-300&.983&2&\scriptsize{.025}&\scriptsize{.859}&\scriptsize{1.0}&.979&3&\scriptsize{.002}&\scriptsize{.964}&\scriptsize{.983}\\
\hline\hline
	\multirow{6}{*}{\rotatebox[origin=c]{90}{\parbox[c]{1cm}{\centering Data Group.}}}
&BaseLine-$n$-1500&.975&5&\scriptsize{.022}&\scriptsize{.884}&\scriptsize{1.0}&.962&5&\scriptsize{.004}&\scriptsize{.949}&\scriptsize{.969}\\\cline{2-12}
&BaseLine-$n$-6500&.975&6&\scriptsize{.02}&\scriptsize{.929}&\scriptsize{1.0}&.965&1&\scriptsize{.003}&\scriptsize{.956}&\scriptsize{.969}\\\cline{2-12}
&IndividualGroup-IG-$n$-5&.978&2&\scriptsize{.018}&\scriptsize{.931}&\scriptsize{1.0}&.961&6&\scriptsize{.004}&\scriptsize{.942}&\scriptsize{.969}\\\cline{2-12}
&IndividualGroup-IG-$n$-21&.977&4&\scriptsize{.019}&\scriptsize{.922}&\scriptsize{1.0}&.964&2&\scriptsize{.003}&\scriptsize{.949}&\scriptsize{.969}\\\cline{2-12}
&SingleSa.Group-SG-$n$-1500&.979&1&\scriptsize{.018}&\scriptsize{.897}&\scriptsize{1.0}&.962&4&\scriptsize{.003}&\scriptsize{.95}&\scriptsize{.966}\\\cline{2-12}
&SingleSa.Group-SG-$n$-6500&.978&3&\scriptsize{.02}&\scriptsize{.915}&\scriptsize{1.0}&.963&3&\scriptsize{.004}&\scriptsize{.952}&\scriptsize{.968}\\\hline
	\end{tabular}		
	\medskip
	\caption{Results for shaping and data grouping}	\label{tab:res}	
\end{table*}


\noindent\textbf{Results}: Table \ref{tab:res} shows the results with a ranking according to the performance on individual data $D_I$ (Acc. on $D_I$) and general data $D_G$ (Acc. on $D_G$). One can observe the trade-off between performance on the two performance metrics, ie. better personalization performance, generally, implies worse general performance. However, two of our \textit{shaping methods} outperform the baseline (optimized towards performance on the general dataset) for both metrics. More precisely, for shaping (upper part in Table  \ref{tab:res}) one of the two parameter settings of each of the three shaping methods outperforms the baseline with respect to personalization performance (p-value $<$ 0.1). More interestingly, the baseline is also worse for the general dataset for two methods. Both ES (p-value $<$ 0.01) and TL(p-value $<$ 0.1) with 400 initial iterations outperforms it. This indicates that initial training with a small dataset can actually improve performance for both metrics. Sample weighing, the best method for personalization, performs significantly worse than the baseline (p-value $<$ 0.001). \\
For \textit{data grouping}, the baseline performs significantly better than all other methods (p-value $<$ 0.1) using $n=6500$ randomly chosen observations as enhanced dataset of an individual.  Also the second and third best method outperform all higher ranked methods with the same p-value. This is intuitive, since these methods train on the most diverse dataset, ie. the enhanced dataset of an individual consists of 5000 samples more than methods with $n=1500$, they are expected to perform better on a general dataset. Diversity also varies depending on how the enhanced dataset is chosen, ie. diversity is largest if data is chosen randomly, followed by choosing all samples of individuals, followed by choosing just the most similar samples. For personalization performance, both baselines are outperformed by the best two data grouping methods (p-value $<$ 0.1).  Thus, for personalization performance it seems highly important to augment the data of the individual only by very similar and not too many samples.


\noindent\textit{Fairness:} Accuracy correlates with fairness. That is to say, overall higher task and personalization performance imply lower standard deviation for individuals. For min-max spread, ie. maximum-minimum, that is sensitive to outliers this holds as well. Thus, overall, one can say that models that perform better should also be preferred from the fairness perspective adopted in this paper though Table \ref{tab:res} also contains a few exceptions. Variation in performance is clearly larger on individual data $D_I$ than on general data $D_G$. This is not surprising, given that the former dataset is of smaller size and, furthermore, discrepancies between individuals are large.


\section{Related Work}  \label{sec:rel}
\textbf{Personalization:} Personalization has been studied in multiple areas, eg.  in cognitive sciences and information systems \cite{fan06}. Personalization using machine learning occurs in contexts such as recommender systems \cite{cheng16,zha18,fus19} or the web \cite{chen04}. Recommender systems implicitly aim at learning features that correlate with users to provide recommendations based on data of the users. As such they are ``inherently'' personal.  Personalization of machine learning itself includes works on interactive machine learning \cite{ame14}, where the goal is to improve a machine learning model in an iterative manner involving humans. 
A conceptualization of personalization in machine learning focusing on explanations without any implementation is given in \cite{sch19}. The idea of providing personalized feedback to a human to generate better inputs to an AI, ie. deep learning, system has also been explored \cite{sch20H2AI}. The training does not necessarily involve data of an individual. Personalization originates from the fact that for a person's input the generated feedback is tailored to the input of that person.
Deep learning with transfer learning was used as one of three methods in \cite{tay17} for multi-task learning with the goal of obtaining personalized models. They show significant gains using personalization -- in particular deep learning. In contrast to our work, for deep learning they discuss just one approach and one evaluation metric, ie. accuracy. While the second approach based on multi-kernel learning (using SVMs)\cite{tay17} has little resemblance to our methods, hierarchical Bayes also allows to group users.\\



\textbf{Goals of personalization:} Goals are diverse \cite{fan06}. They can be stated as architectural (creating a functional and delightful environment with a personalized style), relational (provide a platform for social interaction compatible with privacy needs), instrumental (increase efficiency and productivity), commercial (increase sales and customer loyalty). In our setting, we are given a machine learning task with its objective(s) potentially embracing all of the listed goals of personalization. A service or product serves its original purpose even if personalized, but it is tailored towards the needs of an individual accounting for concerns arising specifically in personalization such as privacy. Elicitation of customer needs is a key objective and challenge in mass personalization \cite{zip01}. Others include customization without overloading the customer with options, production methods that can perform customization on multiple attributes and producing and delivering products for an individual. While for digital products, some aspects such as delivery might bear less relevance, overall these goals are still valid in their abstract form. \\

\noindent\textbf{Goals of machine learning:} Research objectives in machine learning have been categorized into task-oriented studies (improve performance for a given set of tasks), theoretical analysis (mathematical characterization of learning algorithms independent of actual applications) and cognitive simulation (investigation and simulation of human learning)~\cite{mit13}. Here, we focus on task-orientation. Task-oriented metrics are fairly diverse ranging from optimizing performance metrics such as accuracy to minimizing the amount of computation for evaluation (or training), eg. aiming at energy efficiency of models \cite{sch19b}, auto-encoding, eg. data compression or embeddings  \cite{mik13}, on to obtaining explainable models~\cite{sch19}.\\ 

\noindent\textbf{Small sample learning and transfer learning:} Our methodology has roots in several areas of machine learning, ie. small sample learning \cite{shu18}, transfer learning \cite{ben12,lei19} and curriculum learning~\cite{ben09a}. Our work also leverages insights from psychology, ie. conditioning of humans \cite{pet04}. Small sample learning \cite{shu18} deals with the problem of learning a model using a limited amount of data. Common strategies include data augmentation, using knowledge from other domains~\cite{ram17}, transfer learning \cite{ben12}, or prior knowledge on concepts~\cite{ste17}.  Transfer learning \cite{ben12} aims at using existing parts such as features of a trained model for another model aimed at a different problem. Essentially, the idea is to maintain some or all layers of a neural network and fine-tune the network by additional training using a small data set. Other strategies also involve increasing the network capacity (both in terms of number of neurons per layer as well as layers) \cite{wan17}. These techniques are valuable in our context, ie. the (late) shaping approach can be seen as a form of transfer learning (or fine-tuning), since we take an existing network and adjust it for personalization using data of an individual. Weighing of training samples is often done to account for differences in distributions, eg. covariate shift \cite{hua07} for training and test data distributions. 
In contrast, to these works, our objective and setting are different. First, we assume homogeneous data, ie. the data samples of an individual might be contained in the shared global dataset or the global dataset might contain very similar training samples. Thus, compared to classical transfer learning, the expected changes to a model due to transfer learning in our context are more subtle.
While sample reweighing is common to address covariate shift~\cite{hua07}, we do not face covariate shift, as we have knowledge of the distribution of test data. However, ideas from these techniques are still valuable. Second, we do not aim to maximize performance on the data of an individual only, but we strike for a balance between performance on both datasets, though potentially with higher preference for the dataset of an individual. 

Deep learning suffers from ``catastrophic forgetting'': learnt representations are forgotten in the course of continuous learning, eg. \cite{kir17}. This is a particular concern for our work, ie. we might train on the dataset of an individual and later on another general dataset or the other way around. There are attempts to remedy ``catastrophic forgetting'', eg. \cite{kir17}. Approaches such as \cite{kir17} rely on reducing the plasticity of the network by fixing or reducing the tolerable updates to learnt representations. These ideas can also be valuable in our context. However, they have to be applied with great care, because they might conflict with the intention to alter representations based on data of an individual. Transfer learning has also been used for personalization using unlabeled data in the context of activity recognition\cite{fal17}. The idea is that data of a new user is labeled using existing data, eg. data of a similar user. Then, supervised learning approaches can be used.\\

\noindent\textbf{Curriculum learning:} It is known that the ordering of training data influences representations, ie. features, learnt by a neural network \cite{ben09a,gra17,mis17,jia14}. Typically, the goal of curriculum learning \cite{ben09a}, ie. adjusting the ordering  of training data, is to optimize efficiency of the learning process as well as the quality of representations. For example, it can be beneficial to start training using easy examples moving towards more difficult samples while also considering diversity of samples \cite{jia14}.\\

\noindent\textbf{Auto-encoder and GANs for data augmentation:} Variational auto-encoders \cite{hsu17} have been used for data augmentation by adjusting speech of a source domain (with labels) to a target domains (without labels). Auto-encoder with the purpose of denoising have also been used \cite{zhe19} for unsupervised, curriculum learning. Generative adverserial networks(GANs) are known to be useful for data augmentation. In the context of personalization, GANs have also been used to create data similar to that of a particular user. That is, Electrocardiogram (ECG) signals are synthesized for individuals\cite{gol19}. 

\section{Conclusions} \label{sec:conc}
We established multiple foundations for personalization of machine learning. This included the clarification of objectives, derivation of suitable metrics for the objectives and scenarios for evaluation. 
In particular, we evaluated and expanded on the ideas of ``shaping'' and ``data grouping'' that allow to obtain personalized deep learning networks. Our evaluation also showed the inherent trade-off between performance on individual and general data. Shaping can outperform a baseline on both data, though best individual performance leads to strongly reduced performance on the general data. Data augmentation by choosing similar samples for each sample of an individual also increases performance of an individual, if the number of chosen samples is not too large.


\bibliography{ref}{}
\bibliographystyle{IEEEtran}

\end{document}